\def\year{2021}\relax
\documentclass[letterpaper]{article} % DO NOT CHANGE THIS
\usepackage{aaai21}  % DO NOT CHANGE THIS
\usepackage{times}  % DO NOT CHANGE THIS
\usepackage{helvet} % DO NOT CHANGE THIS
\usepackage{courier}  % DO NOT CHANGE THIS
\usepackage[hyphens]{url}  % DO NOT CHANGE THIS
\usepackage{graphicx} % DO NOT CHANGE THIS
\def\UrlFont{\rm}  % DO NOT CHANGE THIS
\usepackage{natbib}  % DO NOT CHANGE THIS AND DO NOT ADD ANY OPTIONS TO IT
\usepackage{caption} % DO NOT CHANGE THIS AND DO NOT ADD ANY OPTIONS TO IT
\usepackage{amsmath}
\usepackage{amssymb}
\usepackage{amsthm}
\usepackage{amsfonts}
\usepackage{mathrsfs}
\usepackage{algorithm}
\usepackage{algorithmic}
\newcommand*\diff{\mathop{}\!\mathrm{d}}

\newtheorem{remark}{Remark}

\title{High Dimensional Level Set Estimation with Bayesian Neural Network}
\author{
    Huong Ha\thanks{Correspondence to: huong.ha@rmit.edu.au}, 
    Sunil Gupta, Santu Rana, Svetha Venkatesh
    \\
}
\affiliations{
    Applied Artificial Intelligence Institute (A\textsuperscript{\rm 2}I\textsuperscript{\rm 2}) \\
    Deakin University, Geelong, Australia\\
}
\begin{document}

\def\UrlFont{\em}
\maketitle

\begin{abstract}
Level Set Estimation (LSE) is an important problem with applications in various fields such as material design, biotechnology, machine operational testing, etc. Existing techniques suffer from the scalability issue, that is, these methods do not work well with high dimensional inputs. This paper proposes novel methods to solve the high dimensional LSE problems using Bayesian Neural Networks. In particular, we consider two types of LSE problems: (1) \textit{explicit} LSE problem where the threshold level is a fixed user-specified value, and, (2) \textit{implicit} LSE problem where the threshold level is defined as a percentage of the (unknown) maximum of the objective function. For each problem, we derive the corresponding theoretic information based acquisition function to sample the data points so as to maximally increase the level set accuracy. Furthermore, we also analyse the theoretical time complexity of our proposed acquisition functions, and suggest a practical methodology to efficiently tune the network hyper-parameters to achieve high model accuracy. Numerical experiments on both synthetic and real-world datasets show that our proposed method can achieve better results compared to existing state-of-the-art approaches.
\end{abstract}

\section{Introduction}

In engineering practice, there are numerous problems which require accurately identifying the regions where the value of a black-box expensive function is higher or lower than a given threshold level. A specific example is alloy design where the task is to determine the regions of alloy compositions that have a desired property (e.g. tensile strength) above a certain level. To solve this, the desired property of a number of alloy compositions are measured and the compositions of interest can be estimated from these measurements. As the measurement process is expensive, thus it is required to minimize the total number of measurements while still maintaining accurate identification for the region of interest. Another example is the algorithmic assurance problem, with the goal being to identify the range of inputs where a machine learning model performs as expected. Specifically, before deploying a machine learning model, it is of interest to identify the range of inputs where the deviation between the machine learning model and the ground truth is lower than a user-specified threshold. A traditional approach is to label the data in the applied domain and estimate the region of interest. As the data labelling process is costly, it is desired to accurately identify this region with a minimal number of sampled data. Other applications of the level set estimation problem can be found in the field of environmental monitoring, manufacturing quality control process, biotechnology, .etc \cite{Bryan2005, Gotovos2013, Bogunovic2016, Zanette2019, Iwazaki2019}. 

% A specific example is in the field of environmental monitoring, with the task being to determine the regions of a lake where the levels of a toxic substance are below a threshold value determined by the field experts. To solve this, a number of locations of the lake can be sampled to check for the toxic level and the regions of interest can be estimated from these measurements. 

In the statistics and machine learning literature, this problem is called the Level Set Estimation (LSE) problem. To solve an LSE problem, the general idea is to formulate it as an active learning problem. First, a small number of measurements of the black-box function are taken to construct a training dataset, and a surrogate model is learned from this training dataset to predict the measurements of all the unmeasured points in the domain. Second, an acquisition function is constructed based on the surrogate model to decide which data point in the domain should be next measured. The black-box function is then evaluated at the chosen data point, and the training dataset is updated with the new measurement. This iterative process continues until the measuring budget is depleted. The regions of interest are then estimated using the learned surrogate model.

There have been a number of research works proposing new methods to tackle the LSE problem, however, all of these approaches are limited due to the use of Gaussian process \cite{Rasmussen2006} as the surrogate model. In particular, it is widely known that standard Gaussian process (GP) (i) does not scale well with high dimensional inputs due to its cubic time complexity, and, (ii) has low representational power because of the limited choices of kernels \cite{Krauth2017}. In this work, we address this issue by suggesting the use of Bayesian Neural Network (BNN) as the surrogate model. Bayesian neural network is a special type of neural network that can produce a probability distribution over the weight parameters \cite{Mackay1992, Neal1995}, thus, it can provide both the prediction and prediction uncertainty for any new data point.

We first consider the \textit{explicit} LSE problem where the threshold level is a fixed user-specified value. Using the uncertainty estimates from the BNN, we derive a theoretic information based acquisition function which can sample the data points that help to most accurately classify the data points in the domain into two sets with outputs below and above the threshold level. We next address the \textit{implicit} LSE problem where the threshold level is defined as a percentage of the (unknown) maximum of the objective function. As the threshold level is unknown, the challenge with the \textit{implicit} LSE problem is that we need to sample the data points that help to most accurately (1) learn the implicit threshold level (or the objective function maximum), and, (2) classify the data points with respect to this estimated threshold level. To solve this problem, we unify the two goals into one goal which is to sample the data point for which if its function value is known, the estimated super- and sub-level set are most accurate. We then construct a novel theoretic information based acquisition function that aims to reduce the uncertainty of the estimated super(sub)-level set, thus, helps to maximally increase the classification accuracy. Finally, we analyse the theoretical time complexity of our proposed acquisition functions, and suggest a practical methodology to efficiently tune the BNN hyper-parameters.
\vspace{-0.2cm}
\paragraph{Contributions} The main contributions of our paper can be summarized as follows:
\begin{itemize}
\item
Proposing for the first time the use of BNN to tackle the LSE problems for high dimensional inputs, 
\item
Deriving two novel sampling methodologies to address the \textit{explicit} and \textit{implicit} LSE problems,
\item
Demonstrating the effectiveness of our proposed methods on both synthetic and real-world problems and comparing with state-of-the-art approaches.
\end{itemize}
\vspace{-0.2cm}
\paragraph{Related Work}

There are various research works aiming to tackle the LSE problem. The work in \cite{Bryan2005} addresses the \textit{explicit} LSE problem by suggesting the \textit{Straddle} heuristic that aims to select the data point that is both predicted to be near the threshold level and also uncertain in terms of its predicted function value. The work was then extended in \cite{Bryan2008} to handle the case when the black-box objective function is a composite function. Later, the work in \cite{Gotovos2013} proposed an algorithm that can solve both the \textit{explicit} and \textit{implicit} LSE problems. Recently, the work in \cite{Shekhar2019} proposed a new method for the \textit{explicit} LSE problem that achieves better theoretical guarantee compared to state-of-the-art methods whilst maintaining a reasonable computational complexity. There are also various works aiming to address some compplex settings of the \textit{explicit} LSE problem such as finding the regions when the probability of the function values larger than a threshold \cite{Zanette2019}, or when the inputs are uncertain \cite{Iwazaki2019} or cost dependent \cite{Inatsu2019}. All of these works rely on Gaussian Process as the surrogate model, and therefore, suffer from the problem of scalability. In the experimental results, we will demonstrate that our proposed approaches outperform these works especially on high dimensional problems.

In a closely related context, there are research works in the field of Bayesian Optimization (BO) with the goal of obtaining the argmax of the black-box objective function using a minimal numbers of sampled data \cite{Snoek2012, Shahriari2016}. The algorithm \textit{TRUVAR} \cite{Bogunovic2016} was developed to unify the \textit{explicit} LSE and BO problem. There are also works that use the ideas behind the LSE problem to develop safe BO methodologies \cite{Sui2015, Sui2018}. Finally, there is the work in \cite{Garnett2012} considering the problem of \textit{active search}, with the goal to sample as many points as possible from one of the level set domains.

\section{Problem Statement and Background}

Let us consider an expensive black-box function $f: \mathcal{X} \rightarrow \mathbb{R}$, where $\mathcal{X}$ is a discrete compact subset of $\mathbb{R}^d$. We assume that we only have noisy measurements of $f(x)$, i.e. querying $f$ at the data point $x \in \mathcal{X}$ results in the noisy function value $f(x) + \epsilon$ where the noise $\epsilon$ follows the normal distribution $\mathcal{N}(0, \sigma^2_{\epsilon})$ and is independent from $x$ or $f(x)$. We consider two types of LSE problems: \textit{explicit LSE} and \textit{implicit LSE}:
\begin{itemize}
\item
The \textit{explicit LSE} problem is to classify all the data points in $\mathcal{X}$ into two subsets based on a user-specified threshold level $h \ (h\in \mathbb{R})$, i.e. to classify $\forall x \in \mathcal{X}$ into a super-level set $H= \lbrace x \in \mathcal{X} \mid f(x) > h \rbrace$ and a sub-level set $L = \lbrace x \in \mathcal{X} \mid f(x) \leq h \rbrace$.
\item
The \textit{implicit LSE} problem is to classify all the data points in $\mathcal{X}$ into two subsets based on the relation with the function maximum, i.e. to classify $\forall x \in \mathcal{X}$ into a super-level set $H= \lbrace x \in \mathcal{X} \mid f(x) > l \max\nolimits_{x \in \mathcal{X}} f(x) \rbrace$ and a sub-level set $L = \lbrace x \in \mathcal{X} \mid f(x) \leq l \max\nolimits_{x \in \mathcal{X}} f(x) \rbrace$ where $l\ (0 \leq l \leq 1)$ is a user-specified threshold.
\end{itemize}
Given a budget of $T$ function evaluations, our goal is to design sequential query point selection strategies to achieve the highest classification accuracies for the two LSE problems.
\vspace{-0.2cm}
\paragraph{Bayesian Neural Networks} Bayesian neural networks (BNN) provide a probabilistic interpretation of neural networks by inferring probability distribution of the networks' weights \cite{Mackay1992, Neal1995}. Given a training data $\mathcal{D}_{tr} = \{x_i, y_i\}_{i=1}^N$, a BNN can provide the posterior distribution $p(\omega \vert \mathcal{D}_{tr})$ with $\omega$ being the neural network weights. This special property of BNNs offers a way to compute the uncertainty of the network's prediction. In particular, for each data point $x$, the predicted distribution $p(y \vert x, \mathcal{D}_{tr})$ can be inferred from the posterior distribution $p(\omega \vert \mathcal{D}_{tr})$ using the formula $p(y \vert x, \mathcal{D}_{tr}) = \int p(y \vert x, \omega) p(\omega \vert \mathcal{D}_{tr}) \diff \omega$.

In practice, performing exact inference to obtain $p(\omega \vert \mathcal{D}_{tr})$ is generally intractable, hence we use a variational approximation technique to approximate this posterior. Specifically, we employ the MC-dropout method \cite{Gal2016}. There are two main reasons for this choice. First, it is both scalable and theoretically guaranteed, i.e. it is shown to be equivalent to performing approximate variational inference to find a distribution $q_{\theta}^*(\omega)$ in a tractable family that minimizes the Kullback-Leibler divergence to the true model posterior $p(\omega \vert \mathcal{D}_{tr})$ \cite{Gal2016}. Second, the method can directly provide a reasonable amount of samples $\hat{\omega}_j$ from the approximate posterior $q_{\theta}^*(\omega)$, which is a key ingredient of our proposed sampling strategy.

\section{Level Set Estimation with Bayesian Neural Network}

\subsection{The explicit LSE problem}

One simple solution is to formulate the \textit{explicit} LSE problem as a classification active learning problem that aims to efficiently train a binary classifier to predict whether a data point belongs to the super- or sub-level set. However, this approach is infeasible in the important cases when the threshold $h$ is close to $\min_{x \in \mathcal{X}} f(x)$ or $\max_{x \in \mathcal{X}} f(x)$. In these cases, the data points in the domain $\mathcal{X}$ mostly belong to one single class, thus, any standard classification active learning method results in incorrect classification for the minor class.
%Let us consider two cases: (1) the threshold $h$ is close to $\min_{x \in \mathcal{X}} f(x)$, or (2) the threshold $h$ is close to $\max_{x \in \mathcal{X}} f(x)$. In these cases, the data points in the domain $\mathcal{X}$ mostly belong to one single class, thus, any standard classification active learning method results in incorrect classification for the minor class.

To overcome this issue, we formulate the \textit{explicit} LSE problem as a regression active learning problem, i.e. actively train a network to predict the function value $f(x)$, $\forall x \in \mathcal{X}$. Using the network prediction $\mathcal{A}(x)$, we can approximate the super-level set $H$ and sub-level set $L$ as $\hat{H}= \lbrace x \in \mathcal{X} \mid \mathcal{A}(x) > h \rbrace$ and $\hat{L} = \lbrace x \in \mathcal{X} \mid \mathcal{A}(x) \leq h \rbrace$. However, unlike the standard regression active learning problem that aims to train a network to most accurately predict the function values for all data points, our goal is to train a network to most accurately predict the function values of only those data points that affect the level set classification accuracy. At iteration $t+1$, given the current observed data $\mathcal{D}_t$, using the mutual information \cite{Houlsby2011, Gal2017} as a way to represent uncertainty, we propose the acquisition function as,
\begin{equation} \label{eq-acq-expLSE}
\begin{aligned}[b]
\alpha_{\text{exp}}(x; \mathcal{D}_t) &= \mathbb{I}(I_y; \omega \vert x, \mathcal{D}_t) \\
& = \mathbb{H}(I_y \vert x, \mathcal{D}_t) - \mathbb{E}_{p(\omega \vert \mathcal{D}_t)} \lbrack \mathbb{H} (I_y \vert x, \omega) \rbrack,
\end{aligned}
\end{equation}
where $y$ denotes the predicted function value corresponding to the data point $x$, $I_y$ is equal $1$ if $y >h$ and equal $0$ otherwise, $\mathbb{H}(I_y \vert x, \mathcal{D}_t)$ denotes the entropy of $I_y$ given the data point $x$ and the current observed data $\mathcal{D}_t$, and $\mathbb{H} (I_y \vert x, \omega)$ denotes the entropy of $I_y$ given $x$ and the model weights $\omega$. The value of $\alpha_{\text{exp}}(x; \mathcal{D}_t)$ is highest when the BNN is most uncertain whether $x$ belongs to the super- or sub-level set.

Using the fact that $p(I_y = c \vert x, \mathcal{D}_t) = \int p(I_y = c \vert x, \omega)p(\omega \vert \mathcal{D}_t)\diff \omega$, we can rewrite $\alpha_{\text{exp}}(x; \mathcal{D}_t)$ as,
\begin{equation}
\begin{aligned}[b] \nonumber
& \alpha_{\text{exp}}(x; \mathcal{D}_t) = -\sum\limits_{c \in \lbrace 0,1 \rbrace} \Big( \int p(I_y = c\vert x, \omega, \mathcal{D}_t) p(\omega \vert \mathcal{D}_t) \diff \omega  \\
& \quad \times \log \int p(I_y = c\vert x, \omega, \mathcal{D}_t) p(\omega \vert \mathcal{D}_t) \diff \omega \Big) \\
& \quad + \mathbb{E}_{p(\omega \vert \mathcal{D}_t)} \sum\limits_{c \in \lbrace 0,1 \rbrace } p(I_y = c\vert x, \omega) \log p(I_y = c\vert x, \omega),
\end{aligned}
\end{equation}
where $c \in \lbrace 0, 1 \rbrace$ and $p(I_y = c\vert x, \omega, \mathcal{D}_t)$ is set to be $c$ if $y \vert x, \omega$ is larger than $h$, and $1-c$ vice versa. Finally, approximating the true posterior $p(\omega \vert \mathcal{D}_t)$ as the MC-dropout posterior $q_{\theta}^*(\omega)$, and using the stochastic forward passes $\lbrace \hat{\omega}_j \rbrace_{j=1}^M$ provided by MC-dropout, we can compute $\alpha_{\text{exp}}(x; \mathcal{D}_t)$ as,
%\begin{equation}
\begin{align} \label{eq-acq-expLSE-approx}
\alpha_{\text{exp}}(x; \mathcal{D}_t) &\approx -\sum\limits_{c \in \lbrace 0,1 \rbrace} \Big( \dfrac{1}{M} \sum\limits_M \hat{p}_c^j \Big) \log \Big( \dfrac{1}{M} \sum\limits_M \hat{p}_c^j \Big) \nonumber \\
& \quad \quad + \dfrac{1}{M} \sum\limits_{c,j} \hat{p}_c^j \log \hat{p}_c^j,
\end{align}
%\end{equation}
with $\hat{p}_c^j$ being the probability of input $x$ with model parameters $\hat{\omega}_j \sim q_{\theta}^*(\omega)$ to be in class $c$.

Our proposed algorithm that solves the \textit{\textbf{exp}licit} \textbf{h}igh dimensional \textbf{LSE} problem using BNN, \textbf{ExpHLSE}, is presented in \textbf{Algorithm} \ref{alg:explicit-LSE-BNN}.

\begin{remark}
It is worth noting that instead of using the mutual information, other criteria such as variation ration \cite{Freeman1965} and entropy \cite{Shannon1948} can also be used to construct the acquisition function.
\end{remark}

\begin{algorithm}[tb]
   \caption{\textbf{ExpHLSE}: Explicit High Dimensional LSE via Bayesian Neural Network}
   \label{alg:explicit-LSE-BNN}
\begin{algorithmic}[1]
   \STATE {\bfseries Input:} Function $f$, domain $\mathcal{X}$, threshold $h$, Bayesian neural network $\mathcal{M}$, acquisition function $\alpha_{\text{exp}}(x)$, initial observations $\mathcal{D}_{init}$, evaluation budget $T$.
   \STATE {\bfseries Output:} Estimated super- and sub-level sets $\hat{H}$, $\hat{L}$.
   \STATE Initialize $\mathcal{D}_0 = \mathcal{D}_{init}$, 
   \FOR {$t = 1, 2, \dots, T$}
   \STATE Train the MC-Dropout BNN using $\mathcal{D}_{t-1}$
   \STATE Generate $M$ forward passes $\lbrace \hat{\omega}_j \rbrace_{j=1}^M$ from the BNN
   \STATE Find $x^* = \text{argmax} \ \alpha_{\text{exp}}(x) (x; \mathcal{D}_{t-1})$ using Eq. (\ref{eq-acq-expLSE-approx})
   \STATE Update observations $\mathcal{D}_t = \mathcal{D}_{t-1} \cup \lbrace x^*, f(x^*) \rbrace$
   \ENDFOR
   \STATE Train the BNN using $\mathcal{D}_{T}$
   \STATE Compute the estimated super- and sub-level sets $\hat{H}= \lbrace x \in \mathcal{X} \mid \mathcal{A}(x) > h \rbrace$ and $\hat{L} = \lbrace x \in \mathcal{X} \mid \mathcal{A}(x) \leq h \rbrace$ where $\mathcal{A}(x)$ is the BNN prediction for the data point $x$.
\end{algorithmic}
\end{algorithm}

\subsection{The implicit LSE problem}

In the \textit{implicit} LSE problem, the goal of the sampling process is to select the data point with two goals: 1) learn the unknown threshold level most accurately, and, 2) classify the data points into the super- and sub-level sets corresponding to the estimated threshold level most accurately. To unify these two goals, we aim to sample the data point for which if its function value is known, the estimated super- and sub-level set are most accurate. For $\forall x \in \mathcal{X}$, we denote $\tilde{f}(x)$ as a random variable denoting all the possible values of $f(x)$. We then define a random variable $\tilde{H}(x)$ as,
\begin{equation} \label{eq-Hx}
\tilde{H}(x) = \lbrace x' \in \mathcal{X} \mid \hat{f}_x(x') > l \max\limits_{x' \in \mathcal{X}} \hat{f}_x(x') \rbrace,
\end{equation}
where 
\begin{equation} \nonumber
\hat{f}_x(x') = \begin{cases}
      \mathbb{E}_{\hat{f}(x') \sim p(\hat{f}(x') \vert \mathcal{D}_t)} \hat{f}(x') & \text{if} \ x' \neq x \\
      \tilde{f}(x) & \text{if} \ x' = x
    \end{cases}, \ \forall x' \in \mathcal{X},
\end{equation}
with $\hat{f}(x')$ is a random variable representing the estimated value of $f(x')$. With this definition, for $\forall x \in \mathcal{X}$, $\hat{f}_x(x’)$ represents all the possible functions $f(x’)$ \textit{when the value of f(x) varies}, thus, the random variable $\hat{H}(x)$ represents all the possible super-level sets $H$ \textit{when the value of $f(x)$ varies}. To obtain an accurate estimated super(sub)-level set within a minimal number of sampled data, the goal is to sample the data point $x$ for whom the random variable $\tilde{H}(x)$ is most uncertain. As $\tilde{H}(x)$ is a random variable representing sets of data points, evaluating which $\tilde{H}(x)$ being the most uncertain is intractable. Thus, we simplify by defining a new random variable $\tilde{G}(x)$ representing the cardinality of the random variable $\tilde{H}(x)$, i.e. $\tilde{G}(x) = \vert \tilde{H}(x) \vert$ . Then we can sample the data point $x$ whose the corresponding random variable $\tilde{G}(x)$ is most uncertain. To characterize the uncertainty of the random variable $\tilde{G}(x)$, we compute its mutual information \cite{Houlsby2011, Gal2017}. Thus, the acquisition function can be formulated as,
\begin{equation}
\begin{aligned} \label{eq-acq-impLSE}
\alpha_{\text{imp}}(x; \mathcal{D}_{t}) &= \mathbb{I}\lbrack \tilde{G}(x); \omega \vert x, \mathcal{D}_t \rbrack \\
&= \mathbb{H}\lbrack \tilde{G}(x) \vert x, \mathcal{D}_t \rbrack -\mathbb{E}_{p(\omega \vert \mathcal{D}_t)} \lbrack  \mathbb{H}\lbrack \tilde{G}(x) \vert x, \omega \rbrack \rbrack,
\end{aligned}
\end{equation}
where $\mathbb{I}\lbrack . \rbrack$ is the mutual information operator and $\mathbb{H} \lbrack . \rbrack$ is the entropy operator. The value of $\alpha_{\text{imp}}(x; \mathcal{D}_t)$ is highest when the BNN is most uncertain in estimating the value of $\tilde{G}(x)$.

%This acquisition function can be approximated using $N$ forward stochastic forward passes $\lbrace \hat{\omega}_j \rbrace_{j=1}^N$ from the posterior MC-dropout distribution.

Firstly, we show how to approximate the first term in Eq. (\ref{eq-acq-impLSE}), $\mathbb{H}\lbrack \tilde{G}(x) \vert x, \mathcal{D}_t \rbrack$. Using the entropy formula, $\mathbb{H}\lbrack \tilde{G}(x) \vert x, \mathcal{D}_t \rbrack$ can be computed as,
\begin{equation} \nonumber
\begin{aligned}
& \mathbb{H}\lbrack \tilde{G}(x) \vert x, \mathcal{D}_t \rbrack \\
& = -\sum\nolimits_{q\in\mathcal{Q}(x)} p(\tilde{G}(x) = q \vert x, \mathcal{D}_t) \log p(\tilde{G}(x) = q \vert x, \mathcal{D}_t),
\end{aligned}
\end{equation}
where $\mathcal{Q}(x)$ is a set consisting of all the possible values of $\tilde{G}(x)$ when $f(x)$ varies. Using the property $p(\tilde{G}(x) = q \vert x, \mathcal{D}_t) = \int p(\tilde{G}(x) = q \vert x, \omega)p(\omega \vert \mathcal{D}_t)\diff \omega$ and the stochastic forward passes $\lbrace \hat{\omega}_j \rbrace_{j=1}^M$ generated by MC-dropout, we can approximate $p(\tilde{G}(x) = q \vert x, \mathcal{D}_t)$ as $ 1/M \sum\nolimits_{j=1}^M p(\tilde{G}(x) = q \vert x, \hat{\omega}_j)$. Therefore, the term $\mathbb{H}\lbrack \tilde{G}(x) \vert x, \mathcal{D}_t \rbrack$ can finally be approximated as,
\begin{equation} \label{eq-acq-impLSE-t1}
\begin{aligned}
\mathbb{H}\lbrack \tilde{G}(x) \vert x, \mathcal{D}_t \rbrack \approx & -\sum\limits_{q\in\mathcal{Q}(x)} \Big( \dfrac{1}{M} \sum\nolimits_{j=1}^M p(\tilde{G}(x) = q \vert x, \hat{\omega}_j) \Big) \\
& \quad \times \log \Big( \dfrac{1}{M} \sum\nolimits_{j=1}^M p(\tilde{G}(x) = q \vert x, \hat{\omega}_j) \Big),
\end{aligned}
\end{equation}

Similarly, the second term $\mathbb{E}_{\omega \sim p(\omega \vert \mathcal{D}_t)} \lbrack  \mathbb{H}\lbrack \tilde{G}(x) \vert x, \omega \rbrack \rbrack$ in Eq. (\ref{eq-acq-impLSE}) can be approximated as,
\begin{equation} \label{eq-acq-impLSE-t2}
\begin{aligned}[b]
& \mathbb{E}_{\omega \sim p(\omega \vert \mathcal{D}_t)} \lbrack  \mathbb{H}\lbrack \tilde{G}(x) \vert  x, \omega \rbrack \rbrack \approx \dfrac{1}{M} \sum\nolimits_{j=1}^M \mathbb{H}\lbrack \tilde{G}(x) \vert  x, \hat{\omega}_j) \rbrack \rbrack \\
& \approx \dfrac{1}{M} \sum\limits_{\substack{j=1,\dots,M \\ q\in\mathcal{Q}(x)}} p(\tilde{G}(x) = q \vert x, \hat{\omega}_j)) \log p(\tilde{G}(x) = q \vert x, \hat{\omega}_j)).
\end{aligned}
\end{equation}

%Using the stochastic forward passes $\lbrace \omega_j \rbrace_{j=1}^N$, $\mathcal{Q}(x)$ can be approximated as the set $\lbrace \tilde{G}_j(x) \rbrace_{j=1}^N$ with $\tilde{G}_j(x) = \vert \tilde{H}_j(x) \vert$ and $\tilde{H}_j(x) = \lbrace x' \in \mathcal{X} \vert \hat{f}_{j,x}(x') > l \max\nolimits_{x' \in \mathcal{X}} \hat{f}_{j,x}(x') \rbrace$ where 
%\begin{equation} \nonumber
%\hat{f}_{j,x}(x') = \begin{cases}
%      1/N \sum\nolimits_{i=1}^N \hat{f}_{\omega_i}(x') & \text{if} \ x' \neq x \\
%      \hat{f}_{\omega_j}(x) & \text{if} \ x' = x
%    \end{cases}, \ \forall x' \in \mathcal{X}.
%\end{equation}
%The probability $p(\tilde{G}(x) = q \vert x)$ can then be computed as the probability of $q$ appeared in the set $ \lbrace \tilde{G}_j(x) \rbrace_{j=1}^N$.

Our proposed \textit{\textbf{imp}licit} \textbf{h}igh dimensional \textbf{LSE} algorithm with BNN, \textbf{ImpHLSE}, is presented in \textbf{Algorithm} \ref{alg:implicit-LSE-BNN}.

\begin{algorithm}[tb]
   \caption{\textbf{ImpHLSE}: Implicit High Dimensional LSE via Bayesian Neural Network}
   \label{alg:implicit-LSE-BNN}
\begin{algorithmic}[1]
   \STATE {\bfseries Input:} Function $f$, domain $\mathcal{X}$, threshold ratio $l$, Bayesian neural network $\mathcal{M}$, acquisition functions $\alpha_{\text{imp}}(x)$, initial observations $\mathcal{D}_{init}$, evaluation budget $T$.
   \STATE {\bfseries Output:} Estimated super- and sub-level sets $\hat{H}$ and $\hat{L}$.
   \STATE Initialize $\mathcal{D}_0 = \mathcal{D}_{init}$, 
   \FOR {$t = 1, 2, \dots, T$}
   \STATE Train the MC-Dropout BNN using $\mathcal{D}_{t-1}$
   \STATE Generate $M$ forward passes $\lbrace \hat{\omega}_j \rbrace_{j=1}^M$ from the BNN
   \STATE Compute all the possible values of the random variable $\tilde{G}(x)$ for each data point $x \in \mathcal{X}$ using $\lbrace \hat{\omega}_j \rbrace_{j=1}^M$
   \STATE Compute the probability distribution of $\tilde{G}(x)$
   \STATE Compute $x^* = \text{argmax} \ \alpha_{\text{imp}} (x; \mathcal{D}_{t-1})$ using Eq. (\ref{eq-acq-impLSE})
   \STATE Update observations $\mathcal{D}_t = \mathcal{D}_{t-1} \cup \lbrace x^*, f(x^*) \rbrace$
   \ENDFOR
   \STATE Train the BNN using $\mathcal{D}_{T}$
   \STATE Compute the estimations of the super- and sub-level sets as $\hat{H}= \lbrace x \in \mathcal{X} \mid \mathcal{A}(x) > l \max_{x \in \mathcal{X}} \mathcal{A}(x) \rbrace$ and $\hat{L} = \lbrace x \in \mathcal{X} \mid \mathcal{A}(x) \leq l \max_{x \in \mathcal{X}} \mathcal{A}(x) \rbrace$ where $\mathcal{A}(x)$ is the BNN prediction for the data point $x$.
\end{algorithmic}
\end{algorithm}

\begin{remark}
If there are multiple data points with the same mutual information $\mathbb{I}\lbrack \tilde{G}(x); \omega \vert x, \mathcal{D}_t \rbrack$, we select the data point with the smallest set $\bigcap\nolimits_{j=1}^M \tilde{H}_j (x)$, i.e. the data point that makes the cardinality $\vert \bigcap\nolimits_{j=1}^M \tilde{H}_j (x) \vert$ to be smallest. This is to ensure that the selected data point is also the data point that causes the set $\tilde{H}(x)$ to be most uncertain.
\end{remark}

\begin{remark}
Similar to the \textit{explicit} LSE problem, instead of using the mutual information, other criteria such as variation ration \cite{Freeman1965} and entropy \cite{Shannon1948} can also be used to construct the acquisition function.
\end{remark}

% the superlevel set $H$. For each data point $x$, we then compute the corresponding values of $\vert H \vert$ when we know the label of this data point, and then the acquisition function is defined as the variance of the $\vert H(x) \vert$. Given $N$ stochastic forward passes $\lbrace \omega_j \rbrace_{j=1}^N$, the cardinality of the estimated superlevel set $H(x)$ corresponding to the data point $x$ is then computed as,
%\begin{equation}
%\hat{H}_j(x) = \lbrace x' \in \mathcal{X} \vert \hat{f}(x') > \hat{f}_{x,j,\max} \rbrace,
%\end{equation}
%where $\hat{f}(x') = 1/N \sum\nolimits_{j=1}^N f_j(x')$ where $f_j(x')$ is the estimation of $f(x')$ using the forward pass $\omega_j$, and $\hat{f}_{x,j,\max}$ is computed as $\max(f_j(x), \max\nolimits_{x' \in \mathcal{X}} \hat{f}(x'))$. 
%
%The acquisition function is then computed as the variance of $\hat{H}_j(x)$,
%\begin{equation}
%\alpha_t(x) = \text{Var}(\vert \hat{H}_j(x) \vert).
%\end{equation}

\subsection{BNN Hyper-parameters Tuning}

%The second challenge is that neural networks normally requires a good set of hyper-parameters in order to achieve high prediction accuracy and this tuning process can be time consuming. In this paper, we also develop a hyper-parameter tuning strategy so as to achieve high network prediction accuracy within a reasonable amount of time.

In general, training a neural network requires a variety of hyper-parameters which contribute to their estimation accuracies. Since our proposed methods belong to the active learning scheme, the data keeps growing, and thus the network hyper-parameters need to be updated to avoid issues such as under-fitting, improper parameter optimization. In the sequel, we will denote the hyper-parameters corresponding to the BNN architecture such as the architecture type, number of layers or neurons as major hyper-parameters and other hyper-parameters such as learning rate, drop-out rate as minor hyper-parameters. We then use the incremental neural architecture search (iNAS) method developed in \cite{Geifman2019} to tune the major hyper-parameters and a grid-search method to tune the minor hyper-parameters. In particular, at iteration $t$, we split the current observed data $\mathcal{D}_t$ into a training and a validation set, and initialize the BNN architecture with fixed major and minor hyper-parameters. We then generate a set of potential candidates for the major hyper-parameters provided by iNAS given the initial major hyper-parameters, and a grid for the minor hyper-parameters. We then choose the major and minor hyper-parameters that optimize the mean square error on the validation dataset. The whole current observed data $\mathcal{D}_t$ is then trained again with these optimal hyper-parameters to generate the BNN. Finally, note that when training the BNN, we use the standard loss function for regression problem, i.e. the loss function is the mean square error between the model predicted output and the ground-truth value.

%To speed up the tuning process, we rely on a unique property of active learning is that the train dataset in the next iteration is closely related to the train dataset in the previous iteration. This means we can transfer knowledge on hyper-parameter tuning in the previous iteration to the next iteration. To do so, we use Bayesian Optimization \cite{Snoek2012} to automatically tune network hyper-parameters, and incorporate a transfer learning scheme to transfer knowledge learned. In particular, we use the transfer learning BO methodology developed in \cite{Wistuba2018}.
%
%We first denote $\lbrace D_i \rbrace_{i=1}^{t-1}$ as the series of dataset obtained at all the iterations up to iteration $t-1$. Then the surrogate model used for the BO process at iteration $t$ can be computed as,
%\begin{equation}
%\begin{aligned}
%\mu(x) &= \big(\sum\nolimits_{i=1}^t b_i \mu_i(x) \big) / \big( \sum\nolimits_{i=1}^t b_i \big), \\
%\sigma^{-2}(x) &= \sum_{i=1}^t v_i \sigma_i^{-2}(x),
%\end{aligned}
%\end{equation}
%where $\lbrace b_i \rbrace$ and $\lbrace v_i \rbrace$ are chosen as $v_i = 1/T$ and $b_i=v_i\sigma_i^{-2}(x)$ for $i=1,\dots,t$. The remaining BO process is then conducted as normal BO methods. Besides, we also employ the method presented in (cite) to initialize the BO process to further enhance its performance. That is, we initialize BO using the maximum value found in previous iterations.

\subsection{Time Complexity of the Proposed Algorithms}

\paragraph{The proposed explicit LSE algorithm} The time complexity of the proposed \textbf{ExpHLSE} acquisition fucntion developed in Eqs. (\ref{eq-acq-expLSE}), (\ref{eq-acq-expLSE-approx}) is $\mathcal{O}(M\vert \mathcal{X} \vert)$, where $M$ is the number of stochastic forward passes by MC-dropout, and $\vert \mathcal{X} \vert$ is the cardinality of the function domain $\mathcal{X}$.
\vspace{-0.2cm}
\paragraph{The proposed implicit LSE algorithm} The time complexity of the proposed \textbf{ImpHLSE} acquisition function in Eqs. (\ref{eq-acq-impLSE}), (\ref{eq-acq-impLSE-t1}), (\ref{eq-acq-impLSE-t2}) is $\mathcal{O}(4\vert \mathcal{X} \vert + 4M \vert \mathcal{X} \vert)$ where $M$ is the number of stochastic forward passes by MC-dropout, and $\vert \mathcal{X} \vert$ is the number of instances in $\mathcal{X}$. This time complexity is computed by splitting the computation of the acquisition function $\alpha_{\text{imp}}(x; \mathcal{D}_t)$ into 3 steps. The first step, which is to compute all the possible values of $\tilde{G}(x)$ for $\forall x \in \mathcal{X}$ when $\tilde{f}(x)$ varies, has the time complexity of $\mathcal{O}((M+3) \vert \mathcal{X} \vert)$. The second step, which is to compute the probability distribution of $\tilde{G}(x)$, has the time complexity of $\mathcal{O}((M+1) \vert \mathcal{X} \vert)$. The final step, which is to compute the two entropy terms, which has the time complexity of $\mathcal{O}(2M \vert \mathcal{X} \vert)$.

\section{Experimental Results}

\begin{figure*}[h]
\begin{center}
\includegraphics[width=156mm,height=45mm]{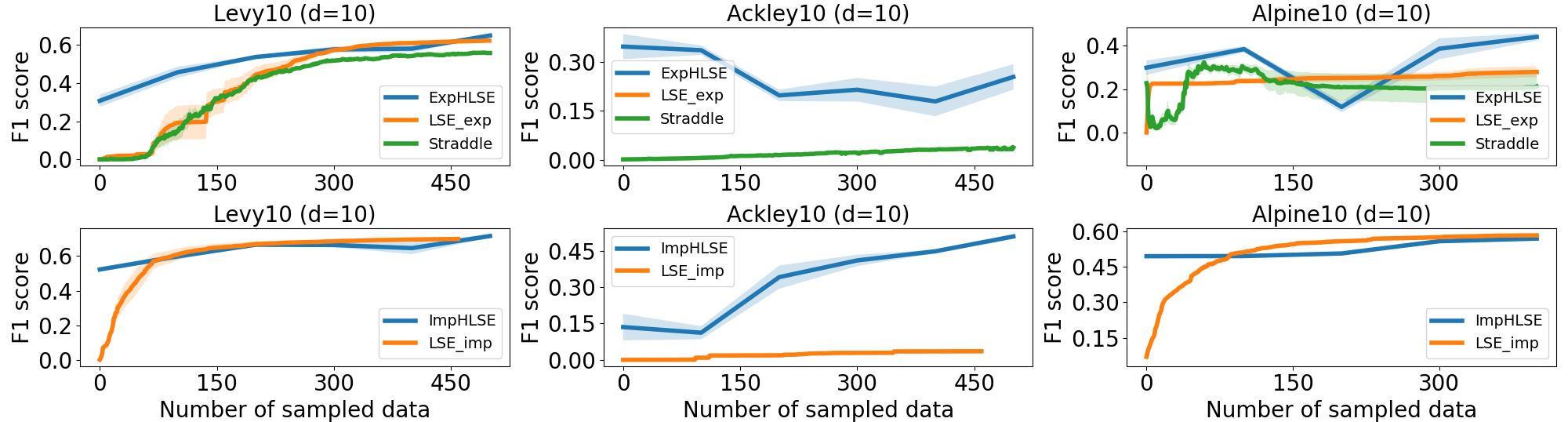}
\caption{Three synthetic benchmark functions: F1-score for super-/sub-level set for \textit{explicit} LSE (top row) and \textit{implicit} LSE (bottom row). Our proposed methods are in blue while baselines are in other colors. Plotting mean and standard error over $5$ repetitions. Method with higher F1-score is better.}
\label{Synthetic_functions}
\end{center}
\end{figure*}

In this section, we evaluate our proposed methods on three benchmark synthetic functions and three real-world problems. The details about the synthetic functions and the real-world problems are in the subsequent sections. For all the problems with dimension $d$, the optimization process is initialized with an initial $3d$ points (for synthetic functions), and $5d$ points (for real-world problems) sampled following a latin hypercube sample scheme \cite{Jones2001}. For all the tasks, the experiments were repeated $5$ times for the synthetic functions and $3$ times for the real-world experiments. All the experiments are running on multiple servers where each server has multiple Tesla V100 SXM2 32GB GPUs.

We compare our proposed methods, \textbf{ExpHLSE} and \textbf{ImpHLSE}, with the state-of-the-art LSE methods. Specifically, for the \textit{explicit} LSE problem, three baselines we compare against are: (1) \textbf{Straddle} in \cite{Bryan2005}; (2) \textbf{LSE}\textsubscript{\textbf{exp}} in \cite{Gotovos2013}; and (3) \textbf{TruVAR} in \cite{Bogunovic2016}. For the \textit{implicit} LSE problem, we compare against the method \textbf{LSE}\textsubscript{\textbf{imp}} in \cite{Gotovos2013} as this is the only method developed to solve the \textit{implicit} LSE problem. For the \textbf{LSE}\textsubscript{\textbf{exp}} and \textbf{LSE}\textsubscript{\textbf{imp}} methods, as suggested in \cite{Gotovos2013}, the exploration-exploitation parameters $\beta_t^{1/2}$ is chosen to be $3$ and the accuracy parameter $\epsilon$ is chosen to increase exponentially from $2\%$ to $20\%$ of the maximum value of each dataset. For the \textbf{TruVAR} method, we use same setting as suggested in \cite{Bogunovic2016}, i.e. we set $\beta_t^{1/2}=\log(\vert \mathcal{X} \vert t^2)$, $\eta_{(1)}=1$, $r=0.1$ and $\bar{\delta}=0$. For GP-based methods, we use the Mat\'{e}rn 5/2 kernel for the GP and we fit the GP using the Maximum Likelihood Estimation method. We also use multi-start to ensure the solution of the optimizer does not stuck in a local minima. For our proposed methods, the BNN we use is a feedforward neural network (FNN). The major hyper-parameters for the iNAS tuning process are the number of layer and the number of neurons per layer whilst the minor hyper-parameters are the learning rate and the drop-out rate. The iNAS tuning process is initialized with a FNN with $1$ layer and $256$ neurons/layer. More details of the hyper-parameter tuning process are in the supplementary material. Finally, for the GP-based methods (except \textbf{TruVAR}), the batch size is set to $1$ and for our proposed BNN based methods and \textbf{TruVAR}, the batch size is set to $10d$ with $d$ being the dimension of the LSE problem. The batch size is the number of data points selected by the acquisition function at each iteration. Setting batch size as $10d$ means we select $10d$ data points with highest acquisition function values. Note that this setting is in favour of the GP-based methods, as the smaller the batch size, the better the performance due to early feedback. Our source code is publicly available at \url{https://github.com/HuongHa12/HighdimLSE}.

\subsection{Synthetic Functions}

%\footnote{Details about these benchmark functions can be found in \textit{https://www.sfu.ca/ \~ ssurjano/optimization.html}}
We evaluate the performance of the methods on three ten dimensional benchmark test functions: Ackley10, Levy10 and Alpine10. As their original function domains are continuous, thus we follow the common practice in \cite{Gotovos2013, Bogunovic2016} which is to randomly select a large number of data points in the continuous domains to generate the corresponding discrete domains $\mathcal{X}$. In particular, for each problem with dimension $d$, we select $10000d$ data points. For the \textit{explicit} LSE problem, for each function, the threshold $h$ is set in such a way that results in the volume of the super-level set being $20\%$ of the domain $X$. For the \textit{implicit} LSE problem, we set the implicit threshold ratio $l$ in such a way that makes $l \max_{x \in \mathcal{X}} f(x)$ to be equal to the threshold $h$ in the \textit{explicit} LSE problem. This is to enable us to compare the two settings. Same with previous works, to measure the effectiveness of the LSE methods, we use the F1-score (i.e., the harmonic mean of precision and recall) with respect to the true super- and sub-level sets. The method resulting in the higher F1-score is the better method. 

In Figure \ref{Synthetic_functions}, we compare the performance of our proposed methods \textbf{ExpHLSE} and \textbf{ImpHLSE} and other baselines. Note that we do not have the performance of \textbf{TruVAR} as its computation time is prohibitive in these cases. Besides, it is also worth noting that the \textbf{LSE}\textsubscript{imp} method has a special stopping condition, thus, it only runs for a limited number of iterations. For the \textit{explicit} LSE problem, it can be seen that \textbf{ExpHLSE} outperforms all baselines on all three functions, especially on the function Ackley10. For the \textit{implicit} LSE problem, our method \textbf{ImpHLSE} also outperforms \textbf{LSE}\textsubscript{imp} significantly on Levy10 and Ackley10, and performs similarly on Alpine10. It is interesting to observe that all the baseline methods perform poorly on the function Ackley10 for both LSE problems. This can be explained by the low representational power of GP, in particular, the number of basic kernels (e.g. RBF, Mat\'{e}rn, polynomial) for GP is limited, and the kernel methods generally encounter the curse of dimensionality issue when learning complex functions.

\subsection{Material Design}

Alloys are produced by mixing many elements to achieve properties that are not possible by a single element. We consider a special alloy that consists of $16$ elements and the sum of all elements is $100\%$. We aim to find the element compositions that achieve the desired properties, that is, when cast a room temperature ($27^o$), these composition results in an Face-Centered-Cubic (FCC) proportion of at least $95\%$. This problem can be formulated as an \textit{explicit} or \textit{implicit} LSE problem for which the explicit threshold level $h$ is set as $0.95$ or the implicit threshold ratio $l$ is set as $0.95$. Same with the current practice in material design, we use the Thermo-Calc software to assist in computing the thermochemical heterogeneous phase equilibria and conducting analysis for evaluating the FCC score of each element composition.

\begin{figure}[h]
\begin{center}
\includegraphics[width=82mm,height=35mm]{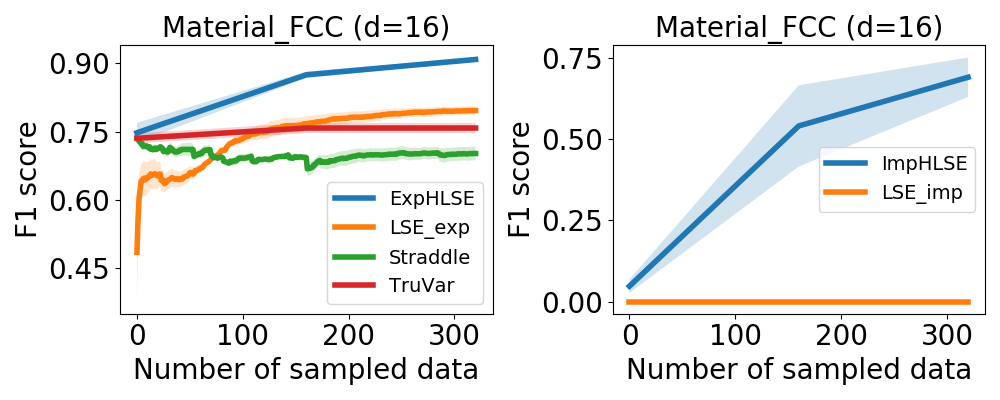}
\caption{Material Design: F1-score for super-/sub-level set for \textit{explicit} LSE (left plot) and \textit{implicit} LSE (right plot). Our proposed methods are in blue while baselines are in other colors. Plotting mean and standard error over $3$ repetitions. Method with higher F1-score is better.}
\label{Material_evaluation}
\end{center}
\end{figure}

In Figure \ref{Material_evaluation}, we evaluate our proposed methods and other baselines using this alloy design problem. For the \textit{explicit} LSE problem (left plot), our proposed method \textbf{ExpHLSE} outperforms other baseline methods significantly. In particular, \textbf{ExpHLSE} can achieve an F1-score of $90\%$ within $300$ sampled data while all the baseline methods can only achieve F1-scores of approximately $65\%-80\%$ within the same number of sampled data. For the \textit{implicit} LSE problem (right plot), our proposed method \textbf{ImpHLSE} also outperforms \textbf{LSE}\textsubscript{imp} by a very high margin. In this particular problem, because of its designed classification methodology, \textbf{LSE}\textsubscript{imp} does not identify any data point belonging to the super-level set, and hence, results in an F1-score of $0$.

% To be more specific, the baseline \textbf{LSE}\textsubscript{imp} has a F1-score of $0$ whilst our proposed method \textbf{ImpHLSE} can reach to an F1-score of $70\%$ when the number of sampled data is less than $300$. 

\subsection{Protein Selection}

\begin{figure}[h]
\begin{center}
\includegraphics[width=84mm,height=34mm]{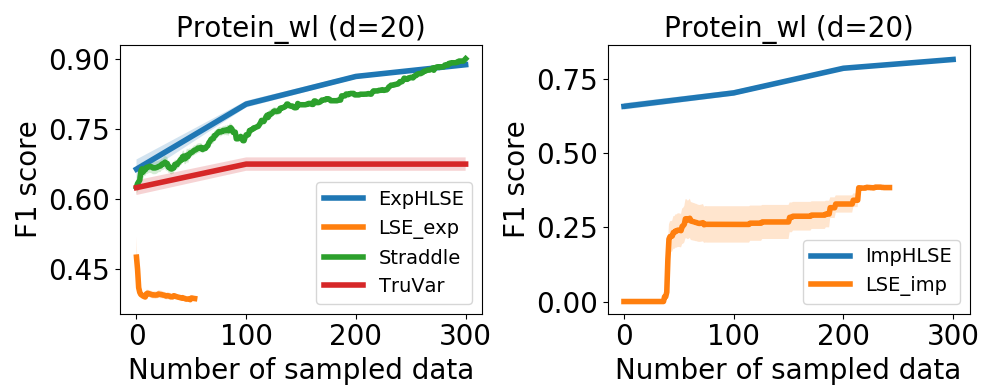}
\caption{Protein Selection: F1-score for super-/sub-level set for \textit{explicit} LSE (left plot) and \textit{implicit} LSE (right plot). Our proposed methods are in blue while baselines are in other colors. Plotting mean and standard error over $3$ repetitions. Method with higher F1-score is better.}
\label{Protein_evaluation}
\end{center}
\end{figure}

In biotechnology, one of the important tasks is to construct new functional proteins by artificially modifying amino acid sequences of proteins \cite{Karasuyama2018, Inatsu2019}. Bio-engineers need to identify the region in the protein space where the protein satisfies the required functional properties. For this task, we use the Rhodopsin-family protein dataset provided in \cite{Karasuyama2018}. This family of proteins are commonly used in optogenetics as it can absorb a light with some certain wavelengths. The goal of this experiment is to estimate the region in the protein feature space where the absorption wavelength is sufficiently large for optogenetics usage. This dataset contains $796$ proteins with each protein having an amino acid sequence vector and a scalar absorption wavelength output. Our goal is to identify the region of protein that (a) has the absorption wavelength larger than $562$ (\textit{explicit} LSE), or, (b) has the absorption wavelength larger than $90.4\%$ of the maximum absorption wavelength (\textit{implicit} LSE).

In Figure \ref{Protein_evaluation}, we compare the performance of our proposed methods \textbf{ExpHLSE} and \textbf{ImpHLSE} with other baselines using this protein selection problem. For the \textit{explicit} LSE problem, our proposed method \textbf{ExpHLSE} outperforms \textbf{TruVAR} and \textbf{LSE}\textsubscript{exp} significantly whilst performs slightly better than \textbf{LSE}\textsubscript{imp}. Note that for this problem, \textbf{LSE}\textsubscript{exp} completes the level set classification process very early (within a small number of sampled data), however, its classification accuracy is low. This can be explained that the GP approximation is not too accurate, and hence, it classifies the data points wrongly and also causes the early termination of \textbf{LSE}\textsubscript{exp}. For the \textit{implicit} LSE problem, \textbf{LSE}\textsubscript{imp} also stops the classification process early and it can achieve a maximum F1-score of $40\%$ whilst our proposed method \textbf{ImpHLSE} can achieve a maximum F1-score of $80\%$ within $300$ sampled data points.

\subsection{Algorithmic Assurance}

We consider the algorithmic assurance problem, which is to find the domain of inputs where a given machine learning model $\mathcal{M}$ will perform as users expected \cite{Sawade2010}. Specifically, given $\mathcal{X}$ being the domain where the machine learning model $\mathcal{M}$ is applied, the problem is to find the domain $L_h = \lbrace x \in \mathcal{X} \mid \vert y_x - \mathcal{M}(x) \vert \leq h \rbrace$ or $L_l = \lbrace x \in \mathcal{X} \mid \vert y_x - \mathcal{M}(x) \vert \leq l \max \vert y_x - \mathcal{M}(x) \vert \rbrace$, where $y_x$ is the true label of the data point $x$, $\mathcal{M}(x)$ is the predicted label by the machine learning model, and $h, l$ are user-specified thresholds. This problem corresponds to the \textit{explicit} and \textit{implicit} LSE problem where the black-box function $f(x)$ is the deviation between the machine learning model output and the ground truth. The goal here is to estimate the domain $L_h$ or $L_l$ using the least number of ground truth sampled from the applied domain $\mathcal{X}$. For the \textit{explicit} LSE problem, the threshold $h$ is set in such a way that results in the volume of the sub-level set $L_h$ being $80\%$ of the domain $X$. For the \textit{implicit} LSE problem, the implicit threshold ratio $l$ is set in such a way that makes $l \max_{x \in \mathcal{X}} f(x)$ to be equal to the threshold $h$ in the \textit{explicit} LSE problem.

For this algorithmic assurance problem, we consider the problem of predicting the performance of highly configurable software systems. We aim to perform an algorithmic assurance analysis on a state-of-the-art machine learning model of predicting software performance, namely DeepPerf \cite{Ha2019}. We want to estimate the sub-level sets where DeepPerf performs as expected on different datasets (i.e. applied domain $\mathcal{X}$). We use two benchmark datasets published in \cite{Siegmund15} for the software performance prediction problem: HSMGP (3456 data points) and HIPACC (13485 data points). The input dimensions of HSMGP and HIPACC are 14, and 33, respectively.

% HSMGP is a highly scalable multi-grid solver for large scale datasets and its performance values are the time to execute  the system on JuQeen, a Blue Gene/Q system located at the Julich Supercomputing Center, Germany. HIPACC is an image processing acceleration framework, which generates efficient low-level code from a high-level specification, and its performance values are the time for solving a test set of partial different equations on an NVIDIA Tesla K20 card. 

\begin{figure}[h]
\begin{center}
\includegraphics[width=80mm,height=66mm]{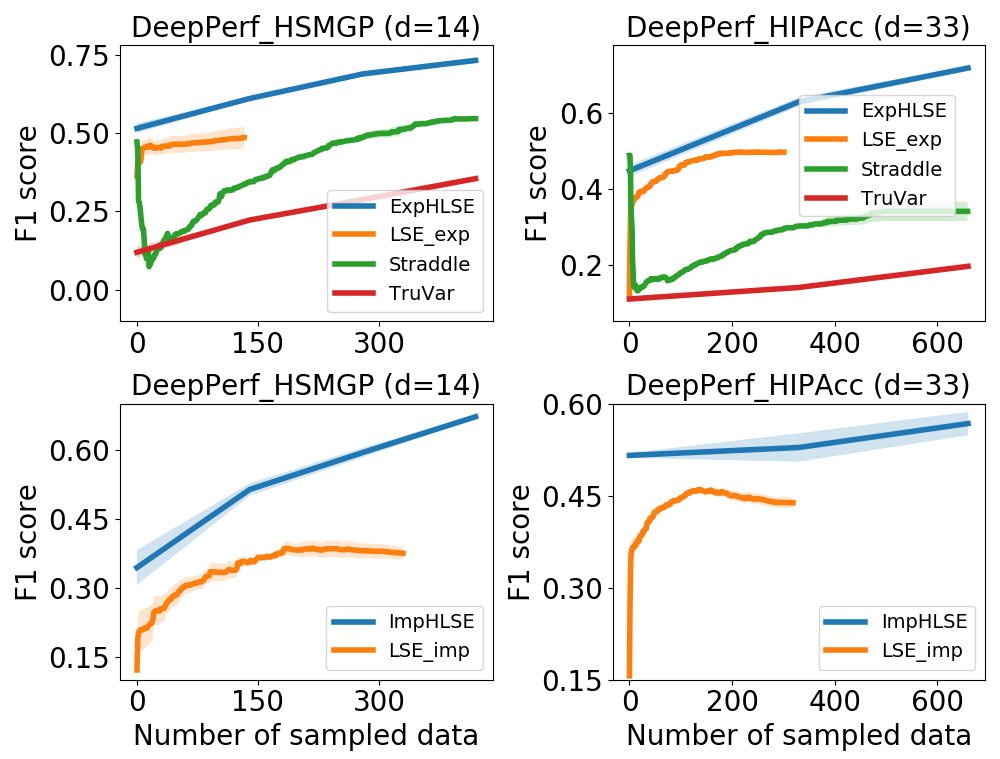}
\caption{Algorithmic Assurance: F1-score for super-/sub-level set for \textit{explicit} LSE (top row) and \textit{implicit} LSE (bottom row). Our proposed methods are in blue while baselines are in other colors. Plotting mean and standard error over $3$ repetitions. Method with higher F1-score is better.}
\label{DeepPerf_evaluation}
\end{center}
\end{figure}

In Figure \ref{DeepPerf_evaluation}, we show the performance of our proposed methods and other baselines. For the \textit{explicit} LSE problem, our method \textbf{ExpHLSE} outperforms all baselines on all datasets. It especially performs better than \textbf{Straddle} and \textbf{TruVar} by a high margin. For the \textit{implicit} LSE problem, our method \textbf{ImpHLSE} also outperforms the state-of-the art \textbf{LSE}\textsubscript{imp} significantly for both datasets.

%\textbf{LSE}\textsubscript{imp} only achieves a maximum F1-score of $45\%$ whilst our method \textbf{ImpHLSE} achieves a maximum F1-score of $60-70\%$ within a limited number of sampled data.

\subsection{Scope and Limitations}

\paragraph{Low dimensional LSE} The proposed BNN approaches might be worse than the GP based methods when the dimension of the LSE problem is low or when the objective function is very simple. We ran the experiments with the functions Branin (d=2), Hartman3 (d=3) and the results show that the GP-based approaches are better than the BNN approaches. Our rule of thumb is that when the input dimension is higher than 10, the BNN-based approaches perform better than the GP-based approaches and vice versa.
\vspace{-0.2cm}
\paragraph{Run time} The run time of the BNN-based approaches, which includes both the hyper-parameter tuning and the BNN training process, is approximately 5-16 hours for each experiment. Specifically, it takes 1-4 hours each time tuning hyper-parameters and several minutes each time training the BNN; and for each experiment, with the batch size of $10d$, we only need to tune and train 3-5 times for the evaluation budget used. On the other hand, the run time of the GP-based methods, which includes the GP hyperparameter tuning and fitting process, is approximately 1-24 hours for each experiment. The slow run time of the GP-based methods is due to two reasons: 1) the batch sizes of the GP-based methods are set as 1 to achieve the best performance, and, 2) the sample budgets are large (as the input is high dimensional) and the GP fitting process is slow when the sampled data is large (GP complexity is cubic in the number of sampled data) \cite{Krauth2017}.

\section{Conclusion}

This paper proposes novel methods to solve the \textit{implicit} and \textit{explicit} LSE problems using Bayesian Neural Networks. Using the uncertainty provided by the BNN, we derive new acquisition functions to sample data points and estimate the super- and sub-level sets in the most efficient manner. We also suggest a practical methodology to efficiently tune the network hyper-parameters to achieve high model accuracy. Numerical experiments on both synthetic and real-world datasets show that our proposed methods can achieve better results compared to state-of-the-art approaches.

\section{ Acknowledgements}

The authors would like to thank Prof. Matthew Barnett and Ms. Manisha Senadeera for the dataset of the material design experiment. This research was partially funded by the Australian Government through the Australian Research Council (ARC). Prof Venkatesh is the recipient of an ARC Australian Laureate Fellowship (FL170100006).

\bibliography{reference}
\bibliographystyle{aaai21}
\end{document}